\documentclass[10pt, a4paper]{article}

\usepackage[final]{lrec2026} 
\usepackage{xcolor}
\usepackage{hyperref}
\usepackage{tikz}
\usetikzlibrary{positioning,calc,fit,shapes,arrows.meta,backgrounds,decorations.pathreplacing,calligraphy}
\usepackage{bm}
\usepackage{xspace}
\usepackage{adjustbox}
\usepackage{graphicx}
\usepackage{siunitx}
\usepackage{fontawesome5}
\usepackage{csquotes}
\usepackage{pgfplots, pgfplotstable}
\pgfplotsset{compat=1.18}
\usepackage{acronym}
\usepackage{numprint}
\usepackage{booktabs}
\usepackage{caption}
\usepackage{subcaption}
\usepackage{pgfplotstable}
\usepackage{progressbar}
\usepackage{amssymb}
\usepackage{amsmath}
\let\OldShowHyphens\showhyphens
\usepackage{listings}
\let\showhyphens\OldShowHyphens
\usepackage{cuted}

\usepackage[true]{anonymous-acm}

\lstdefinelanguage{json}{
  basicstyle=\ttfamily\small,
  numbers=left,
  numberstyle=\tiny,
  stepnumber=1,
  numbersep=5pt,
  showstringspaces=false,
  breaklines=true,
  frame=single,
  backgroundcolor=\color{lightgray},
  literate=
   *{0}{{{\color{numb}0}}}{1}
    {1}{{{\color{numb}1}}}{1}
    {2}{{{\color{numb}2}}}{1}
    {3}{{{\color{numb}3}}}{1}
    {4}{{{\color{numb}4}}}{1}
    {5}{{{\color{numb}5}}}{1}
    {6}{{{\color{numb}6}}}{1}
    {7}{{{\color{numb}7}}}{1}
    {8}{{{\color{numb}8}}}{1}
    {9}{{{\color{numb}9}}}{1}
    {:}{{{\color{punct}{:}}}}{1}
    {,}{{{\color{punct}{,}}}}{1}
    {"}{{{\color{delim}{"}}}}{1},
}

\definecolor{tweetblue}{RGB}{66,133,244}
\definecolor{photored}{RGB}{219,68,55}
\definecolor{imagegreen}{RGB}{15,157,88}

\definecolor{numb}{rgb}{0.7,0.2,0.2}
\definecolor{punct}{rgb}{0,0,0}
\definecolor{delim}{rgb}{0,0,0.5}
\definecolor{lightgray}{rgb}{0.95,0.95,0.95}

\definecolorset{cmyk}{GU-}{}{%
	Goethe-Blau,1,0.2,0,0.4;%
	Hellgrau,0.04,0.04,0.05,0.02;%
	Sandgrau,0.12,0.09,0.13,0;%
	Dunkelgrau,0.25,0.25,0.3,0.75;%
	Purple,0.08,1,0.3,0.36;%
	Emo-Rot,0.04,1,0.8,0.07;%
	Senfgelb,0.01,0.25,1,0.05;%
	Gruen,0.62,0.4,0.87,0.09;%
	Magenta,0.08,0.86,0.12,0.12;%
	Orange,0,0.7,1,0.04;%
	Sonnengelb,0,0.12,0.95,0;%
	Helles-Gruen,0.4,0.17,0.81,0.07;%
	Lichtblau,0.8,0,0.06,0.04}

\definecolor{HUDarkBlue}{RGB}{16,82,132}
\definecolor{HULightBlue}{RGB}{55,140,195} 
\definecolor{HUBlueText}{RGB}{0,115,186}
\definecolor{HUDarkGrey}{RGB}{206,207,209}
\definecolor{HUMedGrey}{RGB}{219,220,221}
\definecolor{HULightGrey}{RGB}{241,241,241} 
\definecolor{HUDarkOrange}{RGB}{146,76,0} 

\colorlet{ColorCount}{GU-Lichtblau}
\colorlet{ColorProcessed}{GU-Gruen}
\colorlet{ColorBase}{GU-Emo-Rot}
\colorlet{ColorRun}{GU-Senfgelb}

\newcommand{\MB}[1]{\textcolor{black}{#1}}
\newcommand{\DB}[1]{\textcolor{black}{#1}}
\newcommand{\Ali}[1]{\textcolor{black}{#1}}
\newcommand{\Giuseppe}[1]{\textcolor{black}{#1}}
\newcommand{\AM}[1]{\textcolor{black}{#1}}
\newcommand{\Mevluet}[1]{\textcolor{black}{#1}}
\newcommand{\Daniel}[1]{\textcolor{black}{#1}}

\newcommand{\AliA}[1]{\textcolor{black}{#1}}

\newcommand{\GerParCor}{\textsc{GerParCor}\xspace}
\newcommand{\DUUI}{\textsc{DUUI}\xspace}
\newcommand{\ParXMulti}{\textsc{Mul\-ti\-Par\-Tweet}\xspace}
\newcommand{\CRAWL}{\textsc{TTLAB\-Tweet\-Craw\-ler}\xspace}

\newcommand{\TextAnnotator}{\textsc{TextAnnotator}\xspace}

\newacro{ddc}[DDC]{Dewey Decimal Classification}

\definecolor{bothcolor}{HTML}{43A047}   
\definecolor{mediacolor}{HTML}{1E88E5}  
\definecolor{textcolor}{HTML}{FB8C00}   
\definecolor{emptycolor}{HTML}{E53935}  

\title{\AM{Extending a Parliamentary Corpus with MPs’ Tweets: Automatic Annotation and Evaluation Using \ParXMulti}}

\name{Mevlüt Bagci, Ali Abusaleh, Daniel Baumartz, Giueseppe Abrami, \\ \fontsize{12pt}{14pt}\selectfont{\textbf{Maxim Konca, Alexander Mehler}}}

\address{Goethe University Frankfurt, Texttechnology \\
         Robert-Mayer-Strasse 10, 60325 Frankfurt am Main \\
         \{bagci, a.abusaleh, baumartz, abrami, konca, mehler\}@em.uni-frankfurt.de\\}

\abstract{
\AliA{
Social media serves as a critical medium in modern politics because it both reflects politicians’ ideologies and facilitates communication with younger generations.
We present \ParXMulti, a multilingual tweet corpus from X that connects politicians’ social media discourse with German political corpus \GerParCor,
thereby enabling comparative analyses between online communication and parliamentary debates. 
\ParXMulti contains \Mevluet{\numprint{39546}} tweets, including \Mevluet{\numprint{19056}} media items.
\AliA{
Furthermore, we enriched the annotation with nine text-based models and one vision-language model (VLM) to annotate \ParXMulti with emotion, sentiment, and topic annotations.
Moreover, the automated annotations are evaluated against a manually annotated subset.
}
\Mevluet{
\ParXMulti can be reconstructed using our tool, \CRAWL, which provides a framework for collecting data from X. 
}
\Mevluet{
To demonstrate
\AliA{ a methodological demonstration,}
we examine whether the models can predict each other using the outputs of the remaining models.
}
In summary, we provide \ParXMulti, a resource integrating automatic text and media-based annotations validated with human annotations, and \CRAWL, a general-purpose X data collection tool.
\Mevluet{Our analysis shows that the models are mutually predictable.}
\Mevluet{In addition, 
\AliA{ VLM-based annotation}
were preferred by human annotators, suggesting that multimodal representations align more with human interpretation.}
}
 \\ \newline \Keywords{Corpora, Tweets, Political Data, Text and Media classification} }

\begin{document}

\maketitleabstract

\section{Introduction} \label{sec:introduction}
Social media platforms such as X, 
TikTok, and Instagram have become increasingly influential in society.
In particular, political content on these platforms has the potential to influence public opinion and even affect electoral outcomes \citep{Olaniran:Williams:2020}.
Social media content is not solely textual; it frequently includes multimedia such as images and videos.  
Consequently, analyses and evaluations of political communication necessitate a corpus that integrates social media content with political datasets.
We address this gap in this study and introduce \ParXMulti, which contains over \numprint{39546} tweets, of which \numprint{19056} include media information.  
\Mevluet{
Over \numprint{15000} of these tweets are from German politicians during the election period from 2024 to 2025, and all are linked to \GerParCor~\citep{Abrami:et:al:2022,Abrami:et:al:2024}.
}
\Mevluet{
We employed our \CRAWL with two methods to download the tweets.
}
The first operates on users, but does not guarantee that the downloaded tweets contain media.  
\Mevluet{
Our second method is search-based and ensures that only tweets including media information are downloaded.  
}
\Mevluet{
Additionally, we used nine text-based models, spanning three types of classification, with \DUUI~\citep{Leonhardt:et:al:2023} to annotate all \numprint{33921} nonempty tweets based on their content.  
}
\Mevluet{
We trained a classifier for each model using annotations generated by the remaining models as input features and optimised the classifiers using evolutionary search\footnote{We use the implementation by \citet{Fortin:et:al:2012}}. 
In addition to the evolutionary search, we used \textit{SHAP} to analyse the impact of each feature input on the prediction model~\citep{Lundberg:Lee:2017}. 
Our evaluation shows that the other models can be used to predict the output classes of a model.
Furthermore, we annotated all three types of classification by using a large language model (LLM) capable of processing media information, where \numprint{18703} tweets were processable.  
}
We manually annotate a sample of \ParXMulti in order to assess and compare the performance of text- and media-based annotation tools.
Our findings indicate that the classification of tweet text should not be conducted in isolation, but rather in conjunction with the relevant media information.
\Daniel{%
Due to data rights, we release tweet IDs in \ParXMulti, but not the tweets or their media information.
All user metadata is included, such as usernames and IDs, as well as the connections to \GerParCor and all automatic and manual annotations.
}
Our code and corpus are available on GitHub under the AGPL license\footnote{The GitHub link will be published after the paper has been accepted}.
\section{Related Work} \label{sec:relatedwork}

\DB{
For over 10 years, researchers have widely used corpora based on tweets from the X platform \citep{McCreadie:et:al:2012,CamachoCollados:Et:al:2022,Blakey:2024}.
Examples include the research on the COVID-19 pandemic, such as the publication of a X corpus spanning from February to March 2020 containing \numprint{90000} tweets by \citet{Naseem:et:al:2021}.
The tweets are automatically annotated with sentiment polarities.
Both the texts and the annotations are available for download.
More corpora are available from \citet{Muller:Salathe:2019,Lopez:Gallemore:2021,Hayawi:et:al:2022,Langguth:et:al:2023}.
}

\DB{%
\citet{Gonzalez:2024} present an annotated, corpus of \numprint{112690} English tweets from 26 news agencies and 27 individuals, from 2009 to 2022.
The results of the automatically generated annotations are visualized in an interactive web application.
However, the authors have not released information on the selected accounts.
They have also not provided the texts of the tweets or their corresponding IDs.
This hinders the reproducibility of the corpus.
\citet{Derczynski:et:al:2016} developed an X Corpus for named entity research that contains approximately \numprint{156000} tokens.
A mixture of experts and crowdsourcing annotated it, and it covers tweets from 2009 to 2014.
The corpus is available for download and includes the texts and annotations.
Additional corpora have been published by \citet{Muhammad:et:al:2016}, including 52 million manually labelled tweets related to 19 different crises, and by \citet{Petrovic:et:al:2010}, including 97 million tweets.
The last corpus is no longer available online.
}

\DB{%
\citet{Barbieri:et:al:2020} proposed a unified X framework to simplify the evaluation and benchmarking of tweet classification tasks.
This framework collects datasets for 7 tasks, including hate speech detection and sentiment analysis.
The tweets are labelled according to the task and based on existing X datasets created by \citet{Mohammad:et:al:2018,Barbieri:et:al:2018,VanHee:et:al:2018,Basile:et:al:2019,Zampieri:et:al:2019,Rosenthal:et:al:2017,Mohammad:et:al:2016}.
}

\DB{%
Although the previous corpora are text-based, research on the use of media content in tweets, such as that of \citet{Thakkar:et:al:2024,Konyspay:et:al:2025,Paul:et:al:2025}, emphasizes the necessity of multimedia corpora.
This paper introduces the X Corpus \ParXMulti, which contains both media and text data.
Similarly, this corpus by \citet{Chen:Zou:2023} is a dataset of \numprint{800000} tweets with AI-generated images that were collected between January 2021 and March 2023.
Although the released data contains generated captions and additional metadata, including the tweet ID, the texts themselves are not available.
From May to July, \citet{Balasubramanian:et:al:2024} built an X corpus by querying tweets related to the 2024 U.S. presidential election.
Tweets, along with additional data such as interaction and media information, are publicly available.
\citet{Grimminger:Klinger:2021} published a corpus of \numprint{3000} tweets from the 2020 US elections that have been manually annotated for stance and hate speech detection.
The release includes the text of the tweets, but not their IDs.
In this paper, we present an X corpus of tweets from German politicians to enable analyses of political communication.
Likewise, \citet{DeSmedt:Jaki:2018} released a corpus of \numprint{125000} tweets and \numprint{4000} linked images.
This corpus is no longer publicly available.
\citet{Weissenbacher:et:al:2024} has released an annotated dataset of 1,250 tweet IDs about German MPs for research on offensive language and hate speech.
\citet{Castanho:Proksch:2022} formed a new corpus by merging the tweets of EU national parliament members \citep{SILVA:PROKSCH:2021} with their parliamentary speech records \citet{Rauh:Schwalbach:2020} from 2018.
Further, however unpublished X corpora from political research include \citet{Conover:et:al:2011,Serrano:et:al:2019,Breeze:2020,Schmidt:et:al:2022,OsamaGhoraba:2023,Hellwig:et:al:2023}.
Our \ParXMulti dataset covers tweets with additional media data from 2024 to 2025
We utilize two methods to link these tweets to \GerParCor dataset.
}

\section{Data} \label{sec:data}
\MB{
\Mevluet{
\ParXMulti focuses on two main aspects of tweets from X.
}
Firstly, it contains tweets from users, paying particular attention to German politicians during the 2025 election period.
This creates an expanded resource of political communication, combining official speeches with social media activity. 
This is an important development, given the significant influence of platforms such as X on contemporary political discourse, particularly during election campaigns \citep{Olaniran:Williams:2020}.
Secondly, \ParXMulti contains a dedicated collection of media content embedded in tweets. 
This provides additional contextual information, thereby enriching the textual data.
To ensure the systematic acquisition of tweets containing only media content, a separate pipeline was developed to specifically target these tweets.
Nevertheless, our media collection enhances the user-based corpus by providing multimodal context for the political communication data.
\Mevluet{
All tweets in \ParXMulti are collected from X using the \textit{X API v2}\footnote{\url{https://docs.x.com/x-api/introduction}}. 
}
\Mevluet{
To facilitate reproducibility, the complete dataset can be reconstructed through our \CRAWL\footnote{All resources will be made publicly available upon acceptance of the paper}.
}
\autoref{fig:workflow} illustrates the full workflow, from data acquisition to the classification of tweets. 
\begin{figure*}[t]
\begin{center}
\begin{adjustbox}{max width=\linewidth}
            \tikzstyle{process} = [rectangle, inner sep=10pt,  minimum height=1cm, text centered, text width=10cm, draw=black, fill=orange!30, font=\huge]
        \tikzstyle{arrow} = [thick,->,>=stealth, line width=5pt]

    \begin{tikzpicture}[node distance=3cm]
        \node (users) [process] {Collect \faTwitter~Usernames};
        \node (userid) [process, below=2cm of users] {Get user id's \faUserFriends~via X API with usernames};
        \node (tweets) [process, below=2cm of userid] {Get tweets \faEdit~via X API of \faUserFriends};
        \node (searchnames) [process, right=4cm of users] {Collect \faSearch~Search Terms};
        \node (mediasearch) [process, below=2cm of searchnames] {Define \faSearch~with has:Media};
        \node (mediatweets) [process, below=2cm of mediasearch] {Get media Tweets \faEdit~via X API of \faSearch};
        \node (media) [process, below=2cm of mediatweets, xshift=-7cm] {\faDownload~\faPhotoVideo~of \faEdit};

        \node (process) [process, right=27cm of users, yshift=-0.39cm] {Process \faPhotoVideo~\&~\faEdit via \DUUI};
        \node (sentiment) [process, below=2cm of process, xshift=8cm] {\faEdit~Sentiment classification};
        \node (emotion) [process, below=2cm of sentiment] {\faEdit~Emotion classification};
         \node (topic) [process, below=2cm of emotion] {\faEdit~Topic classification};
         \node (img) [process, below=2cm of process, xshift=-8cm] {\faFileImage~Sentiment, Emotion and Topic classification LLM};
         \node (video) [process, below=3.625cm of img] {\faFileVideo~Sentiment, Emotion and Topic classification LLM};
        \begin{scope}[on background layer]
            \node[fit=(media) (users) (mediasearch), fill=GU-Goethe-Blau!30!white, draw=GU-Goethe-Blau, line width=2pt, inner sep=20pt, minimum height=14.05cm,label={[yshift=1.5ex]above:{\Huge Download Tweets}}] (download) {};
            \node[fit=(process) (topic) (video), fill=GU-Goethe-Blau!30!white, draw=GU-Goethe-Blau, line width=2pt, inner sep=20pt,label={[yshift=1.5ex]above:{\Huge \DUUI}}] (duui) {};
            \node[fit=(topic) (sentiment), fill=GU-Emo-Rot!30!white, draw=GU-Goethe-Blau, line width=2pt, inner sep=20pt, label={[yshift=1.5ex]above:{\Huge Text \DUUI}}] (textPipe) {};
            \node[fit=(img) (video), fill=GU-Emo-Rot!30!white, draw=GU-Goethe-Blau, line width=2pt, inner sep=20pt, label={[yshift=1.5ex]above:{\Huge Media \DUUI}}] (mediaPipe) {};
        \end{scope}

        \draw [arrow] (users)--(userid);
        \draw [arrow] (userid)--(tweets);
        \draw [arrow] (tweets)|-+(2,-1.75)-|(media);
        
        \draw [arrow] (searchnames)--(mediasearch);
        \draw [arrow] (mediasearch)--(mediatweets);
        \draw [arrow] (mediatweets)|-+(-2,-2.025)-|(media);

        \draw [arrow] (download.east) -- ($(duui.west)$);

        \draw [arrow] (process.south) |- (mediaPipe.east);
        \draw [arrow] (process.south) |- (textPipe.west);
        
    \end{tikzpicture}
\end{adjustbox}
\caption{Building \ParXMulti: Overview of the process of the crawling of tweets and media data and the annotation using \DUUI}
\label{fig:workflow}
\end{center}
\end{figure*}
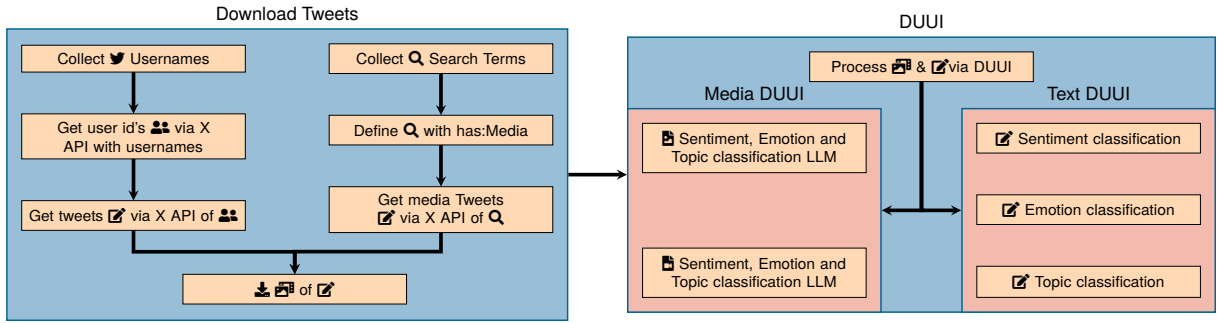
As previously mentioned, our \CRAWL retrieves not only the textual content and standard metadata provided by the X API, but also the associated images and videos.
%
To achieve this, the download process is divided into two steps. 
%
First, tweets are retrieved.
Then, the media information of each tweet is extracted to download the corresponding images or videos. 
%
This paper describes two approaches to collecting tweets, with the second approach ensuring that all downloaded tweets contain media content.
\autoref{fig:media-status} shows the number of tweets that were downloaded, as well as how many of them contain media content in the form of images or videos.
%
For the first collection method, we compiled a list of X usernames belonging to German politicians who held positions in their parties or were officially listed as candidates in the 2025 elections. 
%
All usernames were reviewed manually and corrected where necessary. 
Users without an X account were excluded.
%
We used the X API to retrieve the corresponding user IDs and download all the tweets from their timelines between 6 November 2024 and 23 February 2025.
%
This time frame, which has been chosen, spans from the day after the collapse of the German coalition government until the end of the 2025 elections.
%
We collected a total of 69 tweets from politicians across seven parties.
\autoref{fig:party-user-tweets} shows the number of tweets grouped by party in the left bar chart and the number of tweets from the top 10 individual politicians in the right bar chart.
\Mevluet{
In the \ParXMulti corpus, the \enquote{Die Grünen} fraction has posted the most tweets, and the politician who has posted the most tweets is \enquote{Dagmar Schmidt}.
}
%
The dataset comprises a total of \numprint{15175} tweets from all parties.
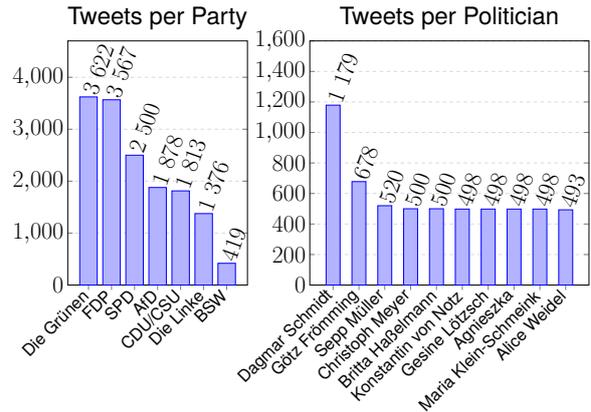
\begin{figure}[t]
\begin{center}
\begin{adjustbox}{max width=\linewidth}
\pgfplotstableread{
	Party	Count
	{Die Grünen}	3622
	{FDP}	3567
	{SPD}	2500
	{AfD}	1878
	{CDU/CSU}	1813
	{Die Linke}	1376
	{BSW}	419
}\datatable

\pgfplotstableread{
	Politician	Party	Count
	{Dagmar Schmidt}	SPD	1179
	{Götz Frömming}	AfD	678
	{Sepp Müller}	CDU/CSU	520
	{Christoph Meyer}	FDP	500
	{Britta Haßelmann}	{Fraktion Bündnis 90/Die Grünen}	500
	{Konstantin von Notz}	{Fraktion Bündnis 90/Die Grünen}	498
	{Gesine Lötzsch}	{Die Linke}	498
	{Agnieszka}	{Fraktion Bündnis 90/Die Grünen}	498
	{Maria Klein-Schmeink}	{Fraktion Bündnis 90/Die Grünen}	498
	{Alice Weidel}	AfD	493
}\Usertable

\begin{tikzpicture}
	
	\begin{axis}[
		ybar,
		bar width=15pt,
		width=7cm,
		height=9cm,,
        xtick pos=bottom,
        xmajorgrids=false,
        ymajorgrids=true,
        grid style={dashed,gray!30},
        ymin=0, ymax=4700,
		symbolic x coords={
			Die Grünen, FDP, SPD, AfD, CDU/CSU, Die Linke, BSW
		},
        yticklabel style={font=\huge},
        xticklabel style={font=\Large, rotate=45, anchor=east},
		xtick=data,
		nodes near coords={
            \rotatebox{90}{\pgfmathprintnumber[1000 sep={~}]{\pgfplotspointmeta}}},
        nodes near coords align={vertical},
        every node near coord/.append style={black, font=\huge, rotate=-20},
		enlarge x limits=0.15,
        title style={font=\huge},
		title={Tweets per Party}
		]
		\addplot table[x=Party,y=Count]{\datatable};
	\end{axis}
	
	\begin{axis}[
		ybar,
		bar width=12pt,
		width=10cm,
		height=9cm,
        ymin=0, ymax=1600,
        xtick pos=bottom,
        xmajorgrids=false,
        ymajorgrids=true,
        grid style={dashed,gray!30},
		symbolic x coords={
			Dagmar Schmidt, Götz Frömming, Sepp Müller, Christoph Meyer, 
			Britta Haßelmann, Konstantin von Notz, Gesine Lötzsch, Agnieszka,
			Maria Klein-Schmeink, Alice Weidel
		},
        yticklabel style={font=\huge},
        xticklabel style={font=\Large, rotate=45, anchor=east},
		xtick=data,
		nodes near coords={
            \rotatebox{90}{\pgfmathprintnumber[1000 sep={~}]{\pgfplotspointmeta}}},
        nodes near coords align={vertical},
        every node near coord/.append style={black, font=\huge, rotate=-20},
		enlarge x limits=0.10,
		title={Tweets per Politician},
        title style={font=\huge},
		at={(8cm,0)}, 
		anchor=origin,
		]
		\addplot table[x=Politician,y=Count]{\Usertable};
	\end{axis}
	
\end{tikzpicture}
\end{adjustbox}
\caption{Number of tweets: party (left); top 10 individual politicians (right) in X dataset}
\label{fig:party-user-tweets}
\end{center}
\end{figure}
}
As previously mentioned, this corpus also extends political speeches of the users by incorporating their social media tweets and associated media content. 
\Giuseppe{
Further information was added to the data set through the selection of the stenographically recorded speeches of the individual politicians from \GerParCor, whereby only speeches from the 20th and 21st legislative periods were included in accordance with the data set. 
The selected speeches do not include any comments from the plenary session, although they include the corresponding speaker and their parliamentary group association.
}
\MB{
\Mevluet{
We employed two methods to establish connections between tweets and \GerParCor. 
%
\Daniel{%
Initially, usernames were directly linked to \GerParCor politicians.
}
}
%
Secondly, we used a cosine similarity approach with a predefined threshold. Speeches that exceeded this threshold were linked to a tweet.
\Mevluet{
To compute the cosine similarity, the speeches of \GerParCor were preprocessed using \textit{spaCy}~\citep{Honnibal:Montani:et:al:2020} to extract the sentences.
}
%
Embeddings were subsequently generated for both the tweets and the sentences from the speeches.
To generate the embeddings of the speeches and tweets, we use the \textit{paraphrase-multilingual-MiniLM-L12-v2} sentence transformer \citep{reimers:gurevych:2019}.
Finally, the cosine similarity was calculated on average for each tweet-speech pair, using the tweet embedding and the corresponding sentence embeddings.
During the downloading process of politicians tweets, the X API provides the capability to retrieve connected tweets.
Additionally, all tweets related to the original tweets (e.g. replies, quotes or retweets) are collected. 
Once the download was complete, we used the URLs in the tweets media information to retrieve any associated images or videos.
%
However, this method does not guarantee that a tweet will contain media information.
In practice, we collected together with the previous method about \numprint{22000} tweets, of which only a little over \numprint{4736} tweets included media content.
%
%
For this reason, we also employed an alternative collection method to ensure the downloaded tweets reliably contained media information.
The second method requires the use of search terms. 
\Mevluet{
To generate these, we use the X API to extract the top 100 hashtags for a predefined list of German- and English-speaking countries\footnote{List of countries and used hashtags are available in the GitHub with the code}.
We assign a weight of 100 to the top-ranked hashtag, 1 to the lowest-ranked hashtag and decrement the weight by one for each subsequent rank.
}
This process is for downloading media-based tweets, not for finding German political tweets. 
For this reason, they are only mapped to \GerParCor via cosine similarity and published in this manner.
The top \numprint{100} hashtags from each country are then combined and sorted by weight.
%
Tweets are retrieved for each search term to ensure that only those containing media links and not retweets are collected.
To this end, the search query is augmented with the filter \texttt{has:media\_link -is:retweet} when using the X API.
Similar to the first method, after downloading the tweets via media links, the corresponding media files are retrieved and stored in base64 format.
\Ali{
\Mevluet{
\ParXMulti contains \numprint{39546} tweets.
}
Among them, \numprint{14756} tweets 
were successfully processed and annotated by both text-based and media-based models. 
A subset of \numprint{3947} tweets
received media-based annotation only, because they did not contain any textual content after the cleaning and filtration stages. 
Conversely, \numprint{19165} tweets 
%
were annotated solely by the text-based model, as they either lacked media or contained broken media links.
Finally, \numprint{1678} tweets
were excluded from our analysis because they contained empty text after cleaning and had no valid media. 
}
Overall total of \numprint{39546} tweets were collected, of which \numprint{19056} contained media. 
Around \numprint{22000} tweets were gathered along with user IDs. 
The remaining \numprint{17474} tweets were obtained through keyword-based searches to ensure that all retrieved tweets contained media content. 
Of these, \numprint{14320} tweets included media information.
%
The discrepancy between the total number of tweets collected and the number of tweets containing media arises because the implementation also downloads tweets connected to the tweets originally retrieved.
The second method therefore ensures that only the original tweets contain media content, rather than the tweets they are connected to.
Nevertheless, the difference between the two approaches is clearly visible.
}
%
%
\begin{figure}[t]
\begin{center}
\begin{adjustbox}{max width=\linewidth} \definecolor{tweetblue}{RGB}{66,133,244}
\definecolor{photored}{RGB}{219,68,55}
\definecolor{imagegreen}{RGB}{15,157,88}

\pgfplotstableread{
Group    Tweets  Images  Videos
Total    39546     14644      4412
User      22072     2464      2272
Search    17474    12180      2140
}\datatable

\begin{tikzpicture}
    \begin{axis}[
        ybar,
        bar width=20pt,
        width=15cm, height=9cm,
        enlarge x limits=0.25,
        ylabel={Number of Tweets},
        ylabel style={font=\huge},
        ticklabel style={font=\huge},
        xticklabel style={font=\huge, rotate=45, anchor=east},
        symbolic x coords={Total, User, Search},
        xtick=data,
        xtick pos=bottom,
        xmajorgrids=false,
        ymajorgrids=true,
        grid style={dashed,gray!30},
        ymin=0, ymax=55500,  
        nodes near coords={
            \rotatebox{90}{\pgfmathprintnumber[1000 sep={~}]{\pgfplotspointmeta}}
        },
        nodes near coords style={
            rotate=-30,
            font=\huge\bfseries,
            /pgf/number format/fixed
        },
        nodes near coords align={vertical},
        every node near coord/.append style={black
        },
        tick style={thick},
        axis line style={thick},
    ]
        \addplot+[ybar, fill=tweetblue!70, draw=none, bar shift=-20pt, rounded corners=2pt] 
            table[x=Group,y=Tweets]{\datatable};

        \addplot+[ybar, fill=photored!70, draw=none, bar shift=0pt, rounded corners=2pt] 
            table[x=Group,y=Images]{\datatable};

        \addplot+[ybar, fill=imagegreen!70, draw=none, bar shift=20pt, rounded corners=2pt] 
            table[x=Group,y=Videos]{\datatable};

    \end{axis}
\end{tikzpicture}
\end{adjustbox}
\caption{\Mevluet{
Numbers of crawled tweets for \ParXMulti via \CRAWL (Tweets: \textcolor{tweetblue}{$\blacksquare$}; Images:~\textcolor{photored}{$\blacksquare$}; Videos: \textcolor{imagegreen}{$\blacksquare$};)
}}
\label{fig:tweet-media}
\end{center}
\end{figure}
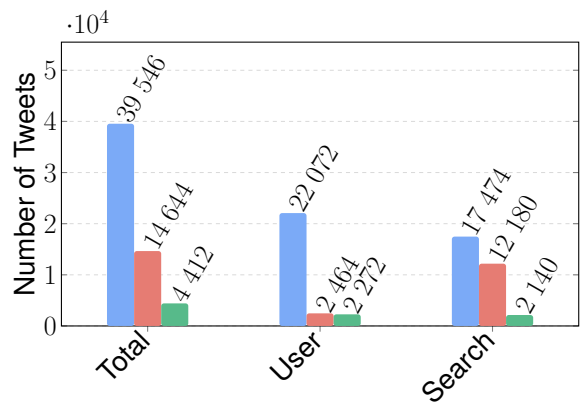
\section{Pipeline}\label{sec:pipeline}
%
\ParXMulti contains text and media content that requires separate processing strategies. Some models are limited to text input, while only a subset can handle multimodal input, used in our pipeline to process images and videos.
\MB{The right side of \autoref{fig:workflow} illustrates the pipeline used to process both text-based and media-based tweets.
All tweets -- regardless of modality -- were processed using \DUUI.
%
%
It also provides direct integration of numerous NLP and LLM-based models for different tasks via Docker containers.
\subsection{Text-based analysis}\label{sec:text-pipeline}
%
\Mevluet{All textual tweets 
were cleaned by removing URLs, hashtags, user mentions, and emojis, and all empty tweets were excluded.}
%
\Mevluet{
After cleaning the initial \numprint{39546} entries, it was found that \numprint{33921} of the tweets contain valid textual information.
}
The remaining tweets were preprocessed using \textit{spaCy} to extract sentence-level segments for classification.}
%
\begin{table}[t]
    \centering
    \begin{tabular}{ll}
        \toprule
        \textbf{Name} & \textbf{Citation} \\
        \midrule
        E1 & \citet{Shivshankar:2023} \\
        E2 & \citet{Bertolini:et:al:2024} \\
        E3 & \citet{Widmann:Wich:2023} \\
        \midrule
        \midrule
        T1  & \citet{Uslu:et:al:2018} \\
        T2 & \citet{Kuzman:Ljubesic:2025:1} \\
        T3 & \citet{Antypas:et:al:2024} \\
        \midrule
        \midrule
        S1 & \citet{Citizenlab:2022} \\
        S2 & \citet{Schmid:2022} \\
        S3 & \citet{Barbieri:et:al:2022} \\
        \midrule
        \midrule
        L1 & \citet{Bai:et:al:2025} \\
        \bottomrule
    \end{tabular}
    \caption{\label{tab:tasksmodels}Overview of Classification Tasks and Used Language Models (E=Emotion, T=Topic, S=Sentiment and L=LLM(Multi Classification)}
\end{table}
\MB{%
\Mevluet{Three classification tasks, emotion, topic and sentiment detection were applied at the sentence level, 
we used three different models for each task.}
The selected models for each task are listed in \autoref{tab:tasksmodels}.
%
\Mevluet{%
The classification output for each tweet was derived by aggregating the predictions at sentence level and calculating the average label representation for each combination of tweet and model.}}
\subsection{Media-based analysis}\label{sec:media-pipeline}
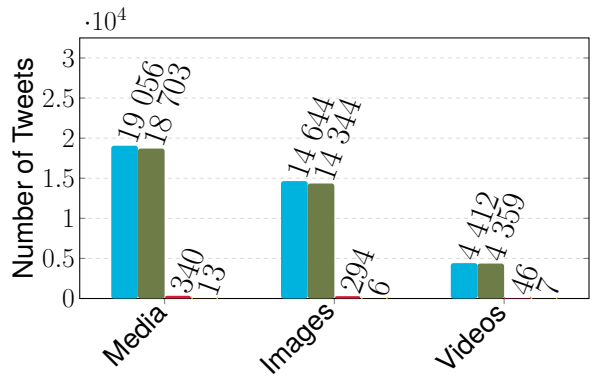
\begin{figure}[t]
    \centering
    \begin{adjustbox}{max width=\linewidth} \pgfplotstableread{
Group	Count	Processed	Base	Runtime
Media	19056	18703	340	13
Images	14644	14344	294	6
Videos	4412	4359	46	7
}\datatable

\begin{tikzpicture}
    \begin{axis}[
        ybar,
        bar width=20pt,
        width=15cm, height=8.5cm,
        enlarge x limits=0.25,
        ylabel style={font=\huge},
        ticklabel style={font=\huge},
        xticklabel style={font=\huge, rotate=45, anchor=east},
        ylabel={Number of Tweets},
        symbolic x coords={Media, Images, Videos},
        xtick=data,
        xtick pos=bottom,
        xmajorgrids=false,
        ymajorgrids=true,
        grid style={dashed,gray!30},
        ymin=0, ymax=32500,      
        nodes near coords={
            \rotatebox{90}{\pgfmathprintnumber[1000 sep={~}]{\pgfplotspointmeta}}
        },
        nodes near coords style={
            rotate=-20,
            font=\Huge\bfseries,
            /pgf/number format/fixed
        },
        nodes near coords align={vertical},
        every node near coord/.append style={black
        },
        tick style={thick},
        axis line style={thick},
    ]
        \addplot+[ybar, fill=ColorCount, draw=none, bar shift=-30pt, rounded corners=2pt] 
            table[x=Group,y=Count]{\datatable};

         \addplot+[ybar, fill=ColorProcessed, draw=none, bar shift=-10pt, rounded corners=2pt] 
            table[x=Group,y=Processed]{\datatable};

         \addplot+[ybar, fill=ColorBase, draw=none, bar shift=10pt, rounded corners=2pt] 
            table[x=Group,y=Base]{\datatable};

        \addplot+[ybar, fill=ColorRun, draw=none, bar shift=30pt, rounded corners=2pt] 
            table[x=Group,y=Runtime]{\datatable};

    \end{axis}
\end{tikzpicture}
    \end{adjustbox}
    \caption{Bar chart summarizing the status of media file processing. Categories include processed, broken, and remaining files for images, videos, and totals \Mevluet{(Tweets: \textcolor{ColorCount}{$\blacksquare$}; Processed: \textcolor{ColorProcessed}{$\blacksquare$}; Base broken: \textcolor{ColorBase}{$\blacksquare$}; Runtime Broken: \textcolor{ColorRun}{$\blacksquare$};)}}
    \label{fig:media-status}
\end{figure}
\MB{
The media data were also processed via \DUUI using the model listed under L1 in \autoref{tab:tasksmodels}.
\Mevluet{
As previously mentioned, \ParXMulti contains a total of \numprint{19056} tweets with media content.
}
}
\Ali{
%
A total of \numprint{18703} media files were successfully processed, consisting of \numprint{14344} image-based tweets and \numprint{4359} video-based tweets.
%
Additionally,  \numprint{340} media tweet files were detected as broken during retrieval, including \numprint{294} images and 46 videos.
%
Furthermore, \numprint{13} media tweet files failed during run-time due to unsupported media formats or incomplete data streams; these include \numprint{6} images and \numprint{7} videos.
%
%
The failure in image processing is primarily attributed to model errors when handling certain image encodings,
particularly malformed base64 strings or image data unsupported by standard decoders.
In contrast, the video failures largely resulted from corrupted or incomplete stream files that could not be parsed.
\autoref{fig:media-status} provides a comparative overview of these statistics, facilitating a clear visual assessment of progress across different media types.
}

\subsubsection{Prompt Construction for Multimodal Media Classification}\label{sec:media-classification}

\Ali{
The prompt is designed to replicate text-based classification behavior within a multimodal reasoning setting, following the \DUUI framework. 
It is instantiated through the L1-model, which serves as the visual-language backbone for multimodal inference. 
The prompt formulation enables reasoning across three interdependent classification dimensions: emotion recognition, sentiment polarity, and topic categorization
\DB{based on the \ac{ddc}~\citep{OCLC:2025}.}
Each subtask is specified using a controlled label set (mapped from the text-based models) to harmonize 
answers.
Emotion recognition uses 5 affective categories (\textit{anger}, \textit{sadness}, \textit{apprehension}, \textit{confusion}, \textit{happiness}) (Model E2 labels); 
sentiment analysis follows a polarity-based scheme (\textit{positive}, \textit{neutral}, \textit{negative}) (Model S1 labels); 
topic classification uses 
the 
\DB{10 main classes in the \ac{ddc} hierarchy,}
which ensures compatibility with knowledge organization standards.
The prompt enforces 3 principles:
\textit{Modularity}: Each subtask is clearly delineated with its own label space and output schema.
\textit{Interpretability}: Predictions must be supported by evidence-based explanations grounded in visual cues.
\textit{Structure}: Output must be formatted as valid JSON with confidence distributions summing to one.
%
These constraints ensure the prompt is both expressive and evaluable. 
\Mevluet{
A high-level abstraction of the full prompt is shown in~\autoref{lst:abstract_prompt}\footnote{The full prompt will be released on GitHub along with the code}. 
}
%
}

\begin{lstlisting}[language=JSON,caption={Abstract-level Prompt for Media-based classification},label={lst:abstract_prompt}]
You are an AI system that performs multimodal understanding of visual input. 
For a given image/video, predict:
1. Emotion (from a fixed label set)
2. Sentiment (positive, neutral, or negative)
3. Topic (using DDC Level 1 classification)
For each, return:
- predicted label
- confidence distribution over possible classes
- explanation grounded in the image
 Ground all outputs in visible content only.
\end{lstlisting}

\subsubsection{\Mevluet{Media-based classification workflow}}

\Ali{
After the prompt is executed, the vLLM model processes the structured input and generates a 
prediction for each subtask as a single output.
These outputs are then refined through a post-processing layer that applies consistency checks, JSON extraction, and semantic validation and aligns the predictions with the controlled label sets.
Finally, the system produces a unified annotation containing emotion, sentiment, and topic information.
}

\section{Manual Annotation}
\label{sec:annotation}

\DB{%
We compare the annotations generated by the two model types (media and text) with a sample of manually annotated data.
%
\Mevluet{
A total of seven professionals manually annotated \numprint{103} tweets (0.26\% of \ParXMulti), resulting in \numprint{721} annotations.
}
The annotation was completed using an annotation web application based on \TextAnnotator~\citep{Abrami:et:al:2020}.
%
In the annotation process, the annotators are shown each tweet, including its text and any accompanying image or video.
%
Following the target classes of the models T1, S1 and E2 as defined in \autoref{sec:media-classification}, they are asked to assign a topic class, a sentiment polarity and an emotion to each tweet.
%
%
%
%
Additionally, the annotation presents the classification results generated by two models, showing the confidence values for each topic, sentiment and emotion class.
The annotators are prompted to select which model produced a better fitting classification:
They can select one of the two models, both, or neither.
We report inter-rater reliability based on Krippendorff's alpha~\citep{Krippendorff:1970} and Fleiss's kappa~\citep{Fleiss:1971}\footnote{We use methods of \citet{Castro:2017} and \citet{Seabold:Perktold:2010} to calculate inter-rater reliability.} in \autoref{tab:annotation_iaa_alpha}.
The results are mixed:
The annotators agree on the topic classification (82\%).
However, the reliability on sentiment and emotion drops to only 45-55\%, suggesting difficulties in assigning these features to tweets -- this has also been reported by e.g. \citet{Bostan:et:al:2020}.
There is less agreement on the question which model produced a more fitting classification distribution with 66\% for topic classes and around 33\% for sentiment and emotion values.
The result show, that the human annotators preferred the results of the media model in all three classifications by a small margin -- 42\% to 34\% in case of the sentiment classification, and 42\% to 33\% for emotion detection.
However, for the topic classification the annotators preferred the media model by a larger margin of 49\% to 23\%. 
%
%
\Mevluet{
We determine the true label based on the manually annotated sample by majority vote (MV) and Dawid-Skene (DS) aggregation\footnote{We use the implementation by \citet{Ustalov:et:al:2024}}~\citep{Dawid:Skene:2025}, and report the Macro F1 scores for the text and media models in \autoref{tab:annotation_fscores}.
}
%
%
%
While the media model achieves a score of around 50\%, the scores for both MV and DS are much lower, at around 10\%.
%
%
Examining the media model, it can be seen that the sentiment and emotion classifications perform similarly to the topic classifications.
%
In contrast to the topic results, the text models perform better for sentiment and emotion.
%
}
\Ali{
Moreover, in order to quantify the quality of the manual annotation, 
we have conducted another annotation approach by randomly assigning labels for the tweets.
We report the same substantial low scores for the same metrics, with almost 0\% for the topic classification, and 29\%, 17\% for the sentiment and emotion classification, respectively. 
These drops in the F1 scores indicate that the original manual annotations are far from random and therefore reliable to a meaningful extent. 
At the same time, the overall performance of the models, particularly in the topic classification task, reveals the intrinsic difficulty of interpreting multimedia content through textual features alone. 
This finding reinforces the assumption that tweets, as highly multimodal entities, require a joint modeling of text and media components to capture their full semantic and affective meaning. 
}

\begin{table}[t]
    \centering
    \begin{tabular}{l|r|r}
        \toprule
         & $\alpha \uparrow$ & $\kappa \uparrow$ \\
        \midrule
         Topic MP & \numprint{0.66} & \numprint{0.66} \\
         Sentiment MP & \numprint{0.33} & \numprint{0.33} \\
         Emotion MP & \numprint{0.34} & \numprint{0.33} \\
         \midrule
         Topic GD & \numprint{0.82} & \numprint{0.82} \\
         Sentiment GD & \numprint{0.54} & \numprint{0.54} \\
         Emotion GD & \numprint{0.47} & \numprint{0.46} \\
        \bottomrule
    \end{tabular}
    \caption{\label{tab:annotation_iaa_alpha}Inter-rater reliability based on Krippendorff's alpha and Fleiss's kappa. MP = Model preference, GD = Generation of gold data}
\end{table}

\begin{table}[t]
    \centering
    \small
    \begin{tabular}{l|l|r|r}
        \toprule
          & Agg. & F1 Media (\#S) & F1 Text (\#S) \\
        \midrule
        Topic & MV & \numprint{0.56} (\numprint{104}) & \numprint{0.11} (\numprint{98}) \\
        Sentiment & MV & \numprint{0.59} (\numprint{104}) & \numprint{0.67} (\numprint{98}) \\
        Emotion & MV & \numprint{0.46} (\numprint{104}) & \numprint{0.56} (\numprint{98}) \\
        \midrule
        Topic & DS & \numprint{0.49} (\numprint{104}) & \numprint{0.13} (\numprint{98}) \\
        Sentiment & DS & \numprint{0.59} (\numprint{104}) & \numprint{0.68} (\numprint{98}) \\
        Emotion & DS & \numprint{0.47} (\numprint{104}) & \numprint{0.56} (\numprint{98}) \\
        \midrule
        Topic & RND & \numprint{0.05} (\numprint{104}) & \numprint{0.09} (\numprint{98}) \\
        Sentiment & RND & \numprint{0.29} (\numprint{104}) & \numprint{0.29} (\numprint{98}) \\
        Emotion & RND & \numprint{0.19} (\numprint{104}) & \numprint{0.17} (\numprint{98}) \\
        \bottomrule
    \end{tabular}
    \caption{\label{tab:annotation_fscores}Validation results using manual annotated samples, \Mevluet{Macro F1 scores}, MV = Majority Voter, DS = Dawid-Skene, RND = Mean of 5 runs with random class assignments, \#S = samples count}
\end{table}

\section{Classification}\label{sec:classification}
In \autoref{sec:text-pipeline}, all cleaned textual tweets ($\mathcal{X}$) were processed using various classification models $\mathcal{T}=\{t_1,\ldots,t_9\}$.
%
\Mevluet{
They generate multiple labels for each tweet, which can then be used to predict the target label of other models.
}
\Mevluet{
The underlying hypothesis H1 is that a list of models can be used to predict the outputs of another model.
}
\Mevluet{
For this purpose, we use, for each model $t_i$, the annotations of the remaining eight models $\mathcal{R}~(t_i\notin \mathcal{R};\; \mathcal{R}\subset\mathcal{T} ;t_i \in \mathcal{T})$ to predict the outputs of $t_i$.
To verify the hypothesis H1, we generate vectors of the model`s annotation outputs for each model $t_i$ and tweet $x$ ($x \in \mathcal{X}$).
For each model, we calculate the mean score for each target category across all sentences in tweet $x$. 
These category scores are then normalised using a softmax function to ensure that the sum of the predicted category scores for each tweet $x$ is equal to one across all target categories.
Using this procedure, we create a vector for each text $x$, based on the remaining $\mathcal{R}$ models corresponding to model $t_i$.
The predicted target label for each model is defined as the category with the highest average probability.
We repeated this procedure for each model and tweet.
}
%
%
Any prediction labels represented by fewer than 50 tweets were removed from the dataset.
\Mevluet{
The dataset was split into two parts: 80\% for training and 20\% for testing. 
We used random forest classifier  to test H1.
}
This study involved a total of nine models, and nine prediction models were trained using this approach.
All models were optimised using an evolutionary search to demonstrate that they can achieve equal or nearly equivalent \Mevluet{Macro F1} scores while utilizing significantly fewer features than previously employed. 
In the evolutionary search, the Random Forest classifier was used as the training model and was trained over 200 epochs.
\begin{figure*}[t]
\begin{center}
\begin{adjustbox}{max width=\linewidth} \colorlet{features}{GU-Goethe-Blau}
\colorlet{sfeatures}{GU-Lichtblau}
\colorlet{f1}{GU-Emo-Rot}
\colorlet{optf1}{GU-Magenta}
\colorlet{labels}{GU-Senfgelb}

\pgfplotstableread{
Model	Features	SFeatures	F1Macro	OptF1Macro	Labels
E1	100.0	59.15	51.8	52.07	6
E2	100.0	66.67	58.73	59.18	5
E3	100.0	62.32	45.94	46.18	8
T1	100.0	61.19	35.17	34.09	10
T2	100.0	70.0	53.73	54.55	16
T3	100.0	56.9	41.75	42.21	18
S1	100.0	34.25	99.26	99.61	3
S2	100.0	41.1	99.26	99.46	3
S3	100.0	38.36	99.26	99.57	3
}\datatable

\begin{tikzpicture}
    \begin{axis}[
        ybar,
        bar width=20pt,
        width=45cm, height=13cm,
        ylabel={Percentage/F1-Score},
        ylabel style={font=\Huge},
        ytick={0,20,40,60,80,100},
        yticklabel style={font=\Huge},
        xticklabel style={font=\Huge},
        symbolic x coords={E1, E2, E3, T1, T2, T3, S1, S2, S3},
        xtick=data,
        xtick pos=bottom,
        xmajorgrids=false,
        ymajorgrids=true,
        grid style={dashed,gray!30},
        ymin=0, ymax=125, 
        nodes near coords={
            \rotatebox{90}{\pgfmathprintnumber[1000 sep={~}]{\pgfplotspointmeta}}
        },
        nodes near coords align={vertical},
        every node near coord/.append style={black, font=\Huge},
        tick style={thick},
        axis line style={thick},
    ]
        \addplot+[ybar, fill=features, draw=none, bar shift=-40pt, rounded corners=2pt] 
            table[x=Model,y=Features]{\datatable};

        \addplot+[ybar, fill=sfeatures, draw=none, bar shift=-20pt, rounded corners=2pt] 
            table[x=Model,y=SFeatures]{\datatable};

        \addplot+[ybar, fill=f1, draw=none, bar shift=0pt, rounded corners=2pt] 
            table[x=Model,y=F1Macro]{\datatable};

        \addplot+[ybar, fill=optf1, draw=none, bar shift=20pt, rounded corners=2pt] 
            table[x=Model,y=OptF1Macro]{\datatable};

         \addplot+[ybar, fill=labels, draw=none, bar shift=40pt, rounded corners=2pt] 
            table[x=Model,y=Labels]{\datatable};

    \end{axis}
\end{tikzpicture}
\end{adjustbox}
\caption{Performance results (\Mevluet{Macro F1} scores and number of features) of the 3~classification tasks using an evolutionary search for optimization of feature selection \Mevluet{(Features: \textcolor{GU-Goethe-Blau}{$\blacksquare$}; Selected Features: \textcolor{GU-Lichtblau}{$\blacksquare$}; Macro F1: \textcolor{GU-Emo-Rot}{$\blacksquare$}; Optimal Macro F1: \textcolor{GU-Magenta}{$\blacksquare$}; Labels: \textcolor{GU-Senfgelb}{$\blacksquare$})}}
\label{fig:classification-statistic}
\end{center}
\end{figure*}
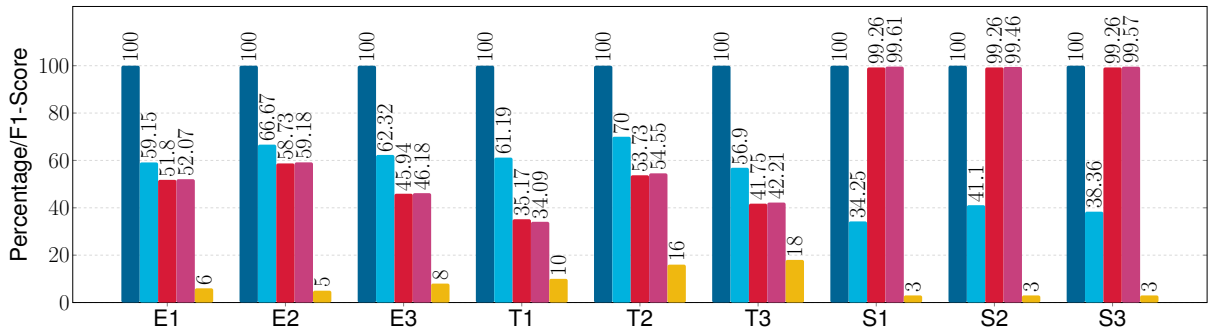
\autoref{fig:classification-statistic} illustrates the \Mevluet{Macro F1} scores before feature selection, using all available feature parameters, and after training the models with the selected feature subset.
Also the number of target categories for each model.
\Mevluet{
As shown in the figure, almost half of the features were not needed to achieve the optimal Macro F1 score.
}
On average, the models utilised around 55\% of the available feature parameters.
The average \Mevluet{Macro F1} score across all models is 65\%, ranging from a minimum of 34\% to a maximum of 99\%.
The \Mevluet{Macro F1} score is nearly the same for all sentiment classification types, at 99\%.
%
This consistency can be attributed to the small number of labels that need to be predicted.
%
\Mevluet{
Additionally, Spearman's rank correlation analysis revealed a strong negative correlation between the Macro F1 score and the number of labels to be predicted (r = -0.85, p < 0.01), suggesting that performance decreases as the number of labels increases.
}
%
\Mevluet{
The distribution of the three labels was more balanced across all sentiment classification models when compared to the topic and emotion models.
}
%
\Mevluet{
For example, the 'joy' label in model E1 comprised just over 50 tweets for training and testing purposes combined.
This made training the model for this label more challenging.
}
\begin{figure*}[t]
\begin{center}
\begin{adjustbox}{max width=0.59\linewidth}
\includegraphics[width=\textwidth]{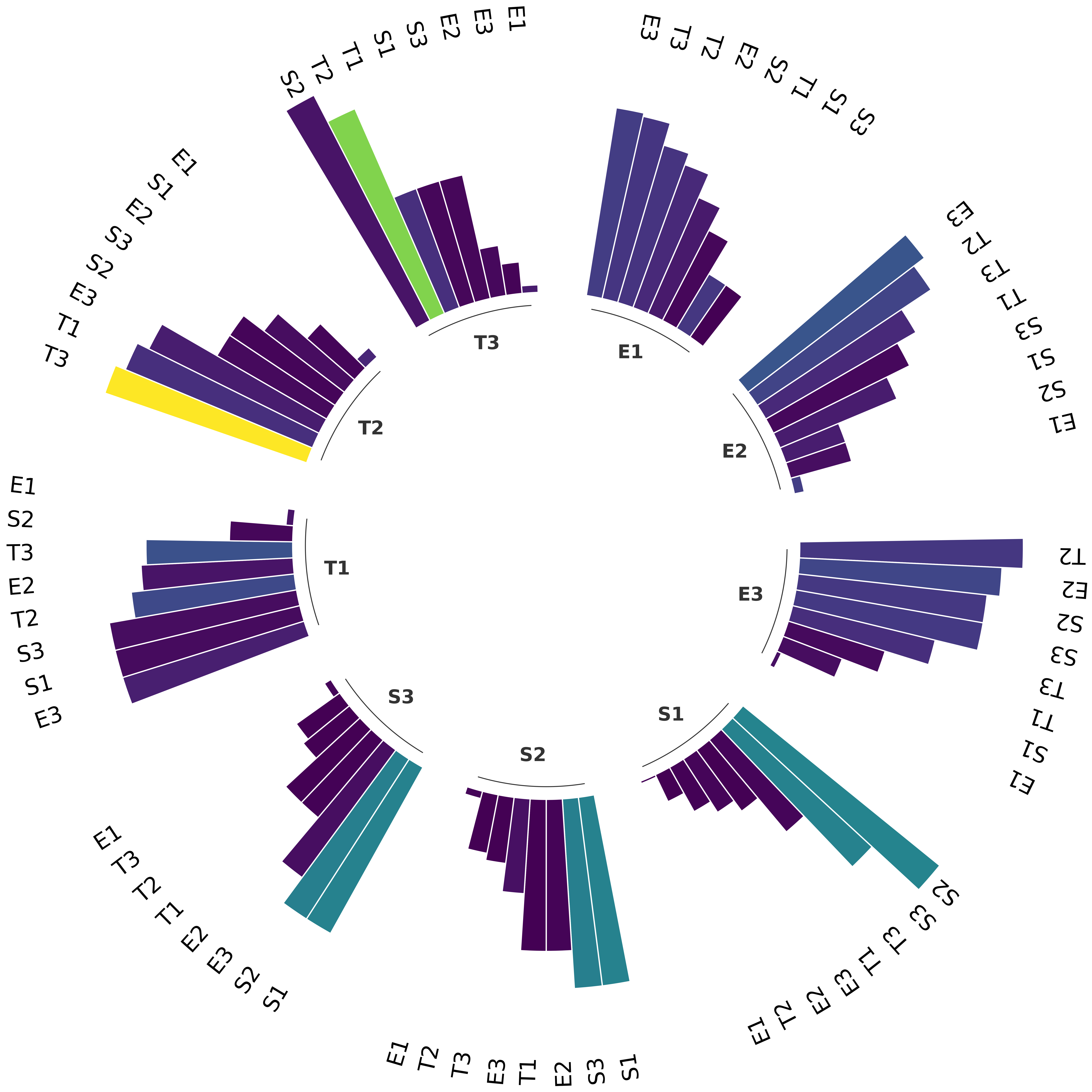}
\end{adjustbox}
\caption{\textit{SHAP} analysis of feature importance in relation to the number of selected features, using the model that performed best in the evolutionary search (shown as a percentage) \Mevluet{\textit{SHAP} impact colouring
\includegraphics[height=1.5ex, width=20.0ex]{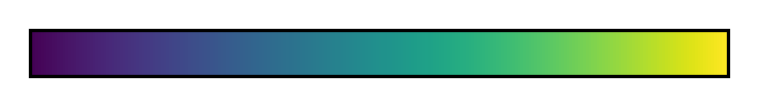} (low: blue; high: yellow)
}}
\label{fig:circular-barchart}
\end{center}
\end{figure*}
%
\Mevluet{
\autoref{fig:circular-barchart} illustrates the proportion of features selected from each feature model by each trained classification model.
}
%
\Mevluet{
A higher bar indicates that a greater proportion of features from that model were selected.
}
%
%
%
\Mevluet{
The E1 model always appears in last place, suggesting that its feature parameters are not useful for any other models following feature selection.
With an average rank of 3.25, the E3 model is the top performer.
Next is the S2 model, with an average rank of 3.75.
}
%
On average, all remaining models occupy middle-ranking positions.
The colours in \autoref{fig:circular-barchart} represent the mean \textit{SHAP} influence of each feature model, ranging from low (blue) to high (yellow).
To ensure comparability, each model was evaluated using \textit{SHAP} with the same training and test splits as in the evolutionary search.
Having obtained the \textit{SHAP} values, which quantify the contribution of each feature parameter to each target label, the mean \textit{SHAP} value was computed for each feature across all labels.
These mean values were then summed for all features belonging to the same model, resulting in an aggregated influence score for each model.
This aggregated score corresponds to the value visualised in \autoref{fig:circular-barchart}.
\Mevluet{
Therefore, the sum of all \textit{SHAP} values indicates the total influence of each feature on the prediction model.
}
\autoref{fig:circular-barchart} suggests a correlation between the features selected through evolutionary search and their \textit{SHAP}-derived influence of the features on the model.
%
\Mevluet{
This observation was verified using a Spearman rank correlation analysis, with the proportion of selected features and the aggregated \textit{SHAP} influence values serving as input variables.
}
%
\Mevluet{
The resulting correlation coefficient is 0.36, with a p-value of less than 0.01. 
}
%
\Mevluet{
This indicates a statistically significant, albeit slight, positive monotonic relationship between the frequency with which features are selected and \textit{SHAP} influence.
}
%
\Mevluet{
The overlap between the top 20 individual feature parameters of each model, ranked once by the percentage of times they were selected during evolutionary search and once by their \textit{SHAP} influence, is exactly 40\%. 
}
This overlap further suggests that the evolutionary search process has successfully identified features that meaningfully impact model predictions.
\Mevluet{
The analysis presented in this section demonstrates the validity of \ParXMulti used in this study and its potential as a valuable resource for further research, particularly in the fields of political science and multimodal analysis, as discussed in previous sections.
}

\section{Conclusion}\label{sec:conclusion}
\MB{
\Mevluet{\ParXMulti offers more than just text data:
%
}
%
it comprises \numprint{15175} tweets from German political figures, which are linked to \GerParCor.  
\Mevluet{
\CRAWL makes it easy to extend the corpus using two downloading methods. 
User-based searches yield more text-only tweets, while media-based searches primarily return tweets containing media. 
%
The two processes together produce a corpus of \numprint{39546} tweets. 
Of these, \numprint{19056} include media information and \numprint{18703} are processable.
We used 9 text-based models, which cover 3 types of classification, to analyse all \numprint{33921} tweets containing text after cleaning.}
%
%
%
\Mevluet{The analysis reveals that the three types -- topic, emotion and sentiment -- can be predicted in relation to each other. 
Furthermore, evolutionary feature selection revealed that achieving the same or higher Macro F1 scores does not necessitate the use of all features.}
%
%
\Mevluet{We also presented a method of automatically annotating media-based information, such as videos and images. 
Using a single prompt, we query a VLM to produce the same three types of classification as text-based models.
A small-scale manual evaluation suggests that media-based annotations are more reliable than text-based ones.
}
However, further 
annotation is required because the current manually labelled dataset is limited. 
%
%
\Mevluet{Nevertheless, this study establishes a framework for the analysis of political data in social media in conjunction with German political speeches.
Furthermore, the provided social media corpus can be used to supplement existing datasets or generate new ones that incorporate media information.}
The corpora and software will be released under the AGPL license on GitHub.}

\section{Ethical Considerations}\label{sec:ethical}
\MB{
This study and the developed tweet corpus have been created with attention to ethical considerations. 
The goal of this work is to provide a large-scale corpus of tweets from German political figures and to make the data collection process as straightforward as possible, while also including media information such as images and videos. 
Although the corpus is based on publicly available tweets, some individual posts may contain content that is offensive, sensitive, or potentially harmful. 
Researchers using the corpus should be aware of this when conducting analyses. 
Similarly, the emotion, sentiment, and topic annotations generated by models may include biases or errors, and their outputs should be interpreted with caution. 
It is acknowledged that the processing pipeline could theoretically be applied to restricted or prohibited content; however, the intention of this work is to support responsible and transparent scientific research.
}

\section{Bibliographical References}\label{sec:reference}

\bibliographystyle{lrec2026-natbib}
\bibliography{lrec2026-example}

\end{document}